\documentclass[11pt,a4paper]{article}
\usepackage[dvipsnames]{xcolor}
\usepackage[hyperref]{naaclhlt2019}
\usepackage{times}
\usepackage{latexsym}
\usepackage{microtype}
\usepackage{url}
\usepackage[hang,flushmargin]{footmisc}

\aclfinalcopy % Uncomment this line for the final submission
\def\aclpaperid{2263} %  Enter the acl Paper ID here
\usepackage{soul}
\usepackage{longtable}
\usepackage{multirow}
\usepackage{booktabs}
\usepackage{amsmath}
\usepackage{amsfonts}
\usepackage{adjustbox}
\usepackage[utf8x]{inputenc}
\usepackage{stmaryrd}

\usepackage[titletoc,toc,title]{appendix}

\def\reg{{\rm\ooalign{\hfil
     \raise.07ex\hbox{\scriptsize R}\hfil\crcr\mathhexbox20D}}}
\allowdisplaybreaks
\usepackage{graphics}

\newcommand{\todo}[1]{\textcolor{WildStrawberry}{\bf\small TODO}}

\title{Fluent Translations from Disfluent Speech in \\End-to-End Speech Translation}
\author{Elizabeth Salesky$^1$, Matthias Sperber$^2$, and Alex Waibel$^{1,2}$\\
  $^1$Carnegie Mellon University, Pittsburgh PA, U.S.A.\\
  $^2$Karlsruhe Institute of Technology, Karlsruhe, Germany\\
  {\tt esalesky@cs.cmu.edu} \\}

\begin{document}
\maketitle

\begin{abstract}
Spoken language translation applications for speech suffer due to conversational speech phenomena, particularly the presence of disfluencies.
With the rise of end-to-end speech translation models, processing steps such as disfluency removal that were previously an intermediate step between speech recognition and machine translation need to be incorporated into model architectures.
We use a sequence-to-sequence model to translate from noisy, disfluent speech to fluent text with disfluencies removed using the recently collected `copy-edited' references for the Fisher Spanish-English dataset. 
We are able to directly generate fluent translations and introduce considerations about how to evaluate success on this task.
This work provides a baseline for a new task, the translation of conversational speech with joint removal of disfluencies.
\end{abstract}

%---------------------------------------------
\section{Introduction \& Related Work}
\label{sec:intro}

Spoken language translation (SLT) applications suffer due to conversational speech phenomena, particularly the presence of disfluencies.
In conversational speech, speakers often use disfluencies such as filler words, repetitions, false starts, and corrections which do not naturally occur in text and may not be desired in translation outputs. 
Disfluency recognition and removal has previously been performed as an intermediate step between speech recognition (ASR) and machine translation (MT), to make disfluent ASR output better-matched to typically clean machine translation training data \cite{cho2013crf,cho2014lrec,wang2010disfluency,honal2005spkdisfluencies,zayats2016disfluency}.
With the rise of end-to-end sequence-to-sequence speech translation systems \cite{weiss2017sequence,bansal2018low}, disfluency removal can no longer be handled as an intermediate step between ASR and MT but needs to be incorporated into the model or handled as a post-processing step.

% DATA
Generating fluent translations from disfluent speech may be desired for simultaneous SLT applications where removing disfluencies will improve the application's clarity and usability.
To train end-to-end speech translation 
requires parallel speech and text translations. 
This introduces data considerations not previously relevant with chained ASR+MT models, as different datasets could be used to train ASR and MT components. 
Where aligned speech and translations exist, data is typically clean speech$\shortrightarrow$clean text, as in news or TED talks, or disfluent speech$\shortrightarrow$disfluent translations, as in Fisher or meeting data, where disfluencies were faithfully included in the references for completeness.
While some corpora with labeled disfluencies exist \cite{cho2014lrec,burger2002isl}, only subsets have been translated and/or released.
\citet{salesky2018slt} introduced a set of fluent references\footnote{Data available at: \href{https://github.com/isl-mt/fluent-fisher}{https://github.com/isl-mt/fluent-fisher}} for Fisher Spanish-English, enabling a new task: end-to-end training and evaluation against fluent references.

Previous work on disfluency removal has treated it as a sequence labeling task using word or span-level labels. 
However, in some cases, simply removing disfluencies from an utterance can create ill-formed output.
Further, corpora can have different translation and annotation schemes: for example for Fisher Spanish-English, translated using Mechanical Turk, \citet{salesky2018slt} found 268 unique filler words due to spelling and casing.
Disfluencies can also be context-specific, such as false starts or corrections where a phrase may be `disfluent' due to its surroundings.
To remove disfluencies as a post-processing step would require a separate model trained with appropriate data and disfluency labels, and may lead to ill-formed output. 
By translating directly to fluent target data instead, we aim to  handle these concerns implicitly. 
We present the first results translating directly from disfluent source speech to fluent target text.

%---------------------------------------------
\section{Data}
\label{sec:data}

For our experiments, we use Fisher Spanish speech \cite{ldcfisherspanish} and with two sets of English translations \cite{salesky2018slt,post2013improved}.
The speech dataset comprises telephone conversations between mostly native Spanish speakers recorded in realistic noise conditions. 
The original English translations were collected through crowdsourcing, as described in \citet{post2013improved}. 
Four references were collected for each of the development and test sets, and one for training. 
The training data consists of 819 conversations yielding ${\sim}160$ hours of speech and 150k utterances; the development and test sets are ${\sim}4$k utterances each.
We use only the first of the two development sets (dev, not dev2). 

This data is conversational and disfluent. 
The original translations faithfully maintain and translate phenomena in the Spanish transcripts such as filler words and hesitations, discourse markers (\textit{you know}, \textit{well}, \textit{mm}), repetitions, corrections and false starts, among others. 
\citet{salesky2018slt} introduced a new set of fluent reference translations collected on Mechanical Turk.
They collected two references for each of the development and test sets, and one for the training set. 
Rather than labeling the disfluencies in the original target data, Turkers were asked to rewrite the utterance in a `copy-edited' manner without disfluent phenomena. 
In some cases, simply removing disfluencies would created ill-formed structure in the resulting utterance. 
This scheme instead creates a sentence-level edit allowing for reordering and insertions as necessary to create fluent content, akin instead to monolingual translation or paraphrasing.
Examples of source transcripts and original translations with the fluent counterparts are shown below in Table \ref{example disfluencies}.

\vspace{-0.6em}
\begin{table}[ht]
\setlength\belowcaptionskip{-10pt}
\centering
\fontsize{10pt}{12pt}\selectfont
\setlength\tabcolsep{3pt} % default value: 6pt
\begin{tabular}{ll} %all from dev
SRC & eh, eh, eh, um, yo pienso que es así \\ %L976
ORG & uh, uh, uh, um, i think it's like that \\
FLT & i think it's like that \\ 
\hline %%%%%%%
SRC & también tengo um eh estoy tomando una clase .. \\ %L635
ORG & i also have um eh i'm taking a marketing class .. \\
FLT & i'm also taking a marketing class \\
\hline %%%%%%%
SRC & porque qué va, mja ya te acuerda que .. \\ %L1675
ORG & because what is, mhm do you recall now that .. \\
FLT & do you recall now that .. \\
\hline %%%%%%%
SRC & y entonces am es entonces la universidad donde  \\
    & ~~~yo estoy es university of pennsylvania\\ %
ORG & and so am and so the university where i am it's  \\
    & ~~~the university of pennsylvania \\
FLT & i am at the university of pennsylvania \\
\end{tabular}
\vspace{-0.3em}
\caption{Disfluency examples in Spanish source (SRC), original (ORG) and fluent (FLT) English translations}
\label{example disfluencies}
\vspace{-0.8em}
\end{table}

%---------------------------------------------
\section{Speech-to-Text Model}
\label{sec:model}

Initial work on the Fisher-Spanish dataset used HMM-GMM ASR models linked with phrase-based MT using lattices \cite{post2013improved,kumar2014some}.
More recently, it was shown in \citet{weiss2017sequence} and \citet{bansal2018low} that end-to-end SLT models perform competitively on this task. 
As in \citet{bansal2018low}, we use a sequence-to-sequence architecture inspired by \citeauthor{weiss2017sequence} but modified to train within available resources; specifically, all models may be trained in less than 5 days on one GPU. 
We build an encoder-decoder model with attention in \texttt{xnmt} \citep{neubig2018xnmt} with 512 hidden units throughout. We use a 3-layer BiLSTM encoder.
We do not use the additional convolutional layers from \citeauthor{weiss2017sequence} and \citeauthor{bansal2018low} to reduce temporal resolution, but rather use network-in-network (NiN) projections from previous work in sequence-to-sequence ASR \cite{zhang2017very,sperber2018self} to get the same total $4\times$ downsampling in time. This gives the benefit of added depth with fewer parameters. 
We closely follow the LSTM/NiN encoder used in \citet{sperber2018self} for ASR and use the same training procedure, detailed in Appendix A.

We extract 40-dimensional mel filterbank features with per-speaker mean and variance normalization with Kaldi \cite{povey2011kaldi}.
We did not see significant difference between 40, 40+deltas and 80-dimensional features in initial experiments, similar to \citet{bansal2018low}, who chose 80-dim. 
\citet{weiss2017sequence} used 240-dim features comprising 80-dim filterbanks stacked with deltas and delta-deltas.
We exclude utterances longer than 1500 frames to manage memory requirements.

Like \citet{weiss2017sequence}, we translate to target characters as opposed to words \citep{bansal2018low}.
We also use an MLP-based attention with 1 hidden layer with 128 units and 64-dimensional target embeddings, though we use only 1 decoder hidden layer as opposed to 3 or 4 in these works.
We use input feeding \cite{Luong2015b}.

All models use the same preprocessing as previous work on this dataset: lowercasing and removing punctuation aside from apostrophes.

%---------------------------------------------
\section{Experiments}
\label{sec:exp}

%--------------
\subsection{Experimental Setup}

We focus on the problem of translating directly from noisy speech to clean references without a separate disfluency removal step.
We first demonstrate the efficacy of our models on the original disfluent Fisher Spanish-English task, comparing to the previously reported numbers on the SLT task \cite{weiss2017sequence,bansal2018low}.
We then compare these results with models trained using the collected `clean' target data with disfluencies removed. 
Finally, we look at the mismatched case where we train on disfluent data and evaluate on a cleaned test set; this is a more realistic scenario, as clean training data is difficult to collect, and we cannot expect to have it for each language and use case we encounter.

We evaluate using both BLEU \cite{papineni2002bleu} and METEOR \cite{meteor2014wmt} to compare different aspects of model behavior on our two tasks.\footnote{BLEU scores are 4-gram word-level BLEU computed using \texttt{multi-bleu.pl} from the Moses toolkit \cite{koehn2007moses}. METEOR is computed using the script from http://www.cs.cmu.edu/˜alavie/METEOR/}
%%%
BLEU assesses how well predicted translations match a set of reference translations using modified n-gram precision, weighted by a brevity penalty in place of recall to penalize short hypothesis translations without full coverage. 
The brevity penalty is computed as $e^{(1-r/c)}$, where $r$ is the length of the reference and $c$ the candidate translation. 
For our task of implicitly removing disfluencies during translation, our generated translations should contain much of the same content but with certain tokens removed, creating shorter translations.
When scoring \textit{fluent output} against the \textit{original disfluent references}, then, differences in BLEU score will come from two sources: shorter n-gram matches, and the brevity penalty.
METEOR, on the other hand, can be considered a more `semantic' evaluation metric. 
It uses a harmonic mean of precision and recall, with greater weight given to recall. 
Further, while BLEU uses exact n-gram matches, METEOR also takes into account stem, synonym, and paraphrase matches.
% By default, these are weighted to 1.0, 0.6, 0.8 and 0.6, respectively, to correlate with human judgments. 
For our fluent task, we aim to maintain semantic meaning while removing disfluent tokens. 
Accordingly, when scored against the fluent target references, we hope to see similar METEOR scores between the disfluent models and fluent models. 
Both metrics are used for a holistic view of the problem: METEOR will indicate if meaning is maintained, but not assess disfluency removal, while BLEU changes will indicate whether disfluencies have been removed.

%%%%
We provide both multi-reference and single-reference BLEU and METEOR scores: the original Fisher target data has four reference translations for the dev and test sets, which boosts scores considerably as hypothesis n-grams can match in any of the references. 
The fluent target data has two references, so the single reference scores better enable comparison between the two tasks.

%--------------------------------
\subsection{Results \& Discussion}

Table \ref{lit comparison} shows our results on the original disfluent data with comparisons to  \citet{weiss2017sequence} and  \citet{bansal2018low}.
All results are single task end-to-end speech translation models.
\citeauthor{weiss2017sequence}'s deeper model reaches a BLEU score of 47.3 on \texttt{test} after 2.5 weeks of training.
Our model is more similar in depth to \citet{bansal2018low}, having both made modifications to train on one GPU in $<5$ days (see Section \ref{sec:model}).
While \citeauthor{bansal2018low} use words on the target side to improve convergence time at a slight performance cost, we are able to use characters like \citeauthor{weiss2017sequence} by having a still shallower architecture (2 fewer layers on both the encoder and decoder), giving us approximately the same training time per epoch they observe with words (${\sim}2$ hours).
We converge to a test BLEU of 33.7, 3-4 BLEU improved over \citeauthor{bansal2018low} on dev and test.
This demonstrates our model has reasonable performance on the original data, providing a strong baseline before turning to our targeted task of directly generating fluent translations.

%----------------
\begin{table}[ht]
\centering
\setlength\tabcolsep{3pt} % default value: 6pt
\begin{tabular}{c|cc|cc|cc} \hline
\bf             & \multicolumn{2}{c|}{\citeauthor{weiss2017sequence}} & \multicolumn{2}{c|}{\citeauthor{bansal2018low}} & \multicolumn{2}{c}{Ours} \\ 
\bf Metric & \bf dev & \bf test & \bf dev & \bf test & \bf dev & \bf test \\ \hline
\small{BLEU 4Ref} & 46.5 & 47.3 & 29.5 & 29.4 & 32.4 & 33.7 \\
\small{BLEU 1Ref} &  --  &  --  &  --  &  --  & 19.0 & 19.6 \\
\hline
\small{METEOR 4Ref} & 36.5 &  --  & 28.2 &  --  & 30.0 & 30.9 \\
\small{METEOR 1Ref} &  --  &  --  &  --  &  --  & 25.1 & 26.1 \\
\hline
\end{tabular}
\caption{Single task end-to-end speech translation using \textbf{original disfluent references} to train and evaluate. Comparing  multi-reference scores using all four references (4Ref) vs average single reference score (1Ref).}
\label{lit comparison}
\end{table}
%----------

Table \ref{fluent ref scores} compares performance of speech translation models trained with the fluent target translations to models trained with the original disfluent translations, as scored on the fluent references. 
Comparing the disfluent and fluent models, we see that METEOR scores are almost the same while BLEU scores are lower with the disfluent model.
This is as we would hope: with our fluent model, we want to generate translations that are semantically the same but with disfluencies removed. 
Therefore similar METEOR scores with similar recall (52) on the fluent references are encouraging.
For BLEU, however, the disfluencies generated by the disfluent model break up n-grams in the fluent references, thereby lowering scores. 

%%%
\begin{table}[ht]
\centering
\setlength\tabcolsep{3pt} % default value: 6pt
\begin{tabular}{l|c|cc|cc} \hline
\bf & & \multicolumn{2}{c|}{\bf dev} & \multicolumn{2}{c}{\bf test}  \\ 
\bf Model & \bf Metric & \bf 1Ref & \bf 2Ref & \bf 1Ref & \bf 2Ref \\ \hline
% \blue{Pretrain+Finetune}     & BLEU &  --  &  --  &  --  &  --  \\ %BLEU
Disfluent  & BLEU & 13.0 & 16.2 & 13.5 & 17.0 \\ %BLEU
Fluent     & BLEU & 14.6 & 18.1 & 14.6 & 18.1 \\ %BLEU
\hline
Disfluent  & METEOR & 22.2 & 23.9 & 23.1 & 24.8 \\ %METEOR
Fluent     & METEOR & 22.3 & 24.0 & 23.1 & 24.9 \\ %METEOR 
\hline
\end{tabular}
\caption{End-to-end model performance evaluated with \textbf{new fluent references}. Comparing average single reference scores (1Ref) vs multi-reference scores using both generated references (2Ref).}
\label{fluent ref scores}
\end{table}
%%%

Comparing single-reference scores with Table \ref{lit comparison}, we see that they are distinctly lower. This is to be expected with the shorter fluent references; a difference of a single token carries greater weight. Translating directly to the fluent references is a more challenging task. As shown in Table \ref{example disfluencies}, the original English translations and Spanish speech are very one-to-one while the edited translations introduce deletions and reorderings. In learning to generate fluent translations, the model needs to learn to handle these more inconsistent behaviors.

\begin{figure}[ht]
    \includegraphics[width=1.0\linewidth]{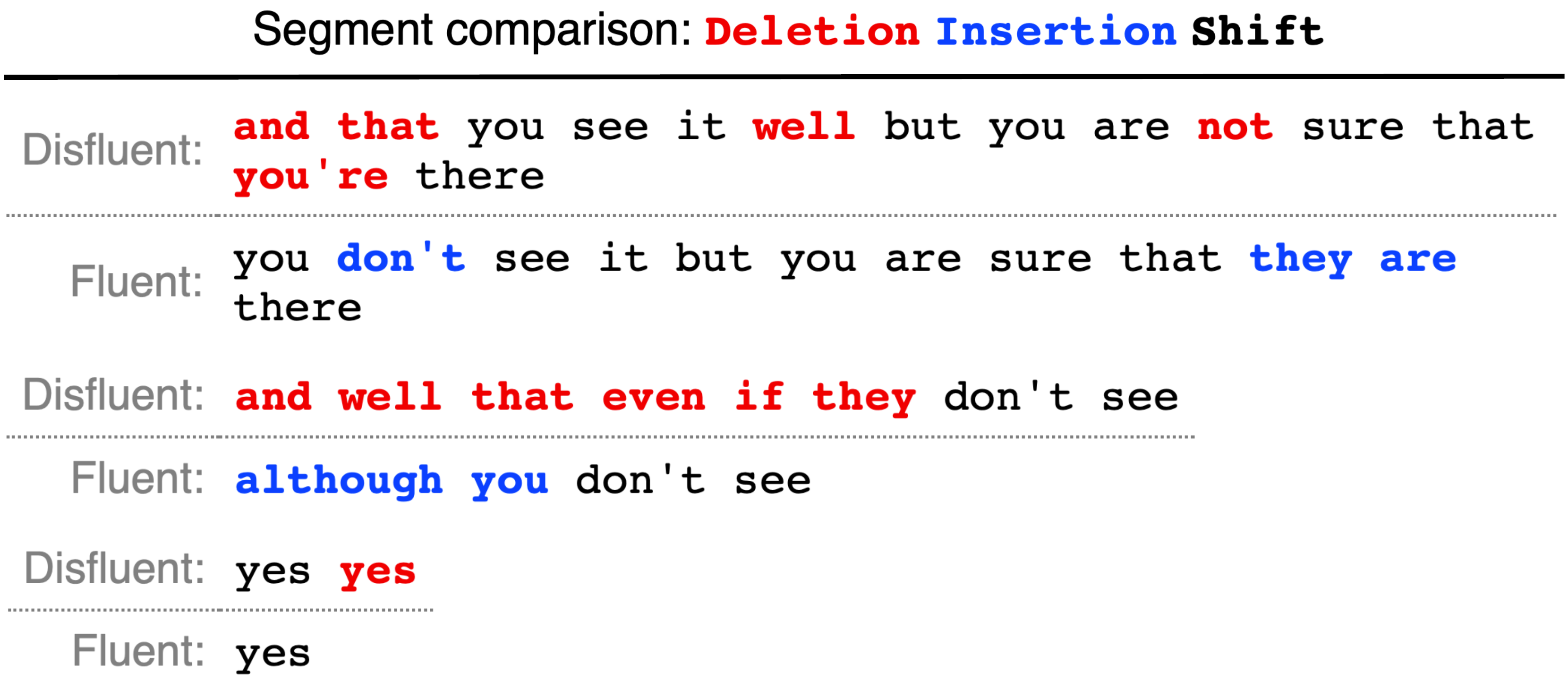}

  \caption{Comparison of example outputs generated by disfluent and fluent models, created with CharCut \cite{charcut2017}.}
  \label{output diff}
\end{figure}

%----------------
Figure \ref{output diff} shows a visual comparison between outputs generated by the two models.
Using the fluent target data to train constrains the model output vocabulary, so filler words such as \textit{`um', `ah', `mhm'} are not generated.
We also see significant reductions in repetitions of both words and phrases from the model trained with fluent reference translations. 
Further, we also see instances where the fluent model generates a shorter paraphrase of a disfluent phrase, as in the 2nd example.

%----------------

Disfluency removal for speech translation has traditionally been done as an intermediate step between ASR and MT to better-match additional clean corpora used for MT training; we do not compare to a pipeline approach here. 
However, to contextualize these results, we compare disfluency removal as a post-processing step after end-to-end speech translation with the original 
%%%
\begin{table}[ht]
\centering
\setlength\tabcolsep{4pt} % default value: 6pt
\begin{tabular}{l|cc|cc} \hline
\bf & \multicolumn{2}{c|}{\bf dev} & \multicolumn{2}{c}{\bf test}  \\ 
\bf Model & \bf 1Ref & \bf 2Ref & \bf 1Ref & \bf 2Ref \\ \hline
% End-to-end Disfluent & 13.0 & 16.2 & 13.5 & 17.0 \\ %Table 3
% End-to-end Fluent    & 14.6 & 18.1 & 14.6 & 18.1 \\ 
% \hline
Postproc. Filter   & 13.6 & 16.5 & 13.5 & 16.8 \\ 
Postproc. MonoMT   & 14.4 & 17.8 & 14.4 & 18.0 \\ 
\hline
\end{tabular}
\caption{End-to-end disfluent model with different post-processing steps. Performance evaluated with \textbf{new fluent references}.} 
% Comparing end-to-end models to post-processing step.}
\label{postproc}
\end{table}
%%%
disfluent parallel data. Simply filtering filler words and repetitions from the disfluent model (Filter) outputs as a post-processing step, the dev scores improve slightly, but test stays the same or decreases.
In some cases, treating disfluency removal as a filtering task can reduce the fluency of an utterance: \\
\vspace{-1.5em}
\begin{table}[ht]
\centering
\setlength\tabcolsep{4pt} % default value: 6pt
\begin{tabular}{ll}
Disfluent & \it mm well and from and the email is a  \\
 & \it scandal the spam.\\ 
Fluent & \it the email is a scandal it's spam. \\
\end{tabular}
\vspace{-0.7em}
\end{table}
\\
\noindent  A filtering or tagging system may not capture all false starts or corrections, leading to lower fluency, and requires labeled spans. 
Treating the post-processing step as a monolingual translation task (MonoMT) rather than a filtering task allows for reordering and insertions, which we saw boost fluency.
We trained a 4-layer BiLSTM encoder-decoder model to translate between the disfluent and fluent English references and applied this to the output of the end-to-end disfluent model. 
BLEU scores approach the results with the end-to-end fluent target model (Table \ref{fluent ref scores}), but we note, this requires the same resources as the direct task.

%----------------
%%%
\begin{table}[b]
\setlength\belowcaptionskip{-5pt}
\centering
\setlength\tabcolsep{4pt} % default value: 6pt
\begin{tabular}{l|c|cc|cc} \hline
\bf & & \multicolumn{2}{c|}{\bf dev} & \multicolumn{2}{c}{\bf test}  \\ 
\bf Model & \bf Metric & \bf 1Ref & \bf 4Ref & \bf 1Ref & \bf 4Ref \\ \hline
Fluent     & BLEU & 16.6 & 29.8 & 17.0 & 30.4 \\ %BLEU
Disfluent  & BLEU & 19.0 & 32.4 & 19.6 & 33.7 \\ %BLEU
\hline
Fluent     & METEOR & 21.8 & 25.9 & 22.7 & 27.0 \\ %METEOR
Disfluent  & METEOR & 25.1 & 30.0 & 26.1 & 30.9 \\ %METEOR
\hline
\end{tabular}
\caption{Performance evaluated with \textbf{original disfluent references}. Comparing average single reference scores (1Ref) vs multi-reference scores using all references (4Ref).}
\label{disfluent ref scores}
\end{table}

Showing the importance of fluent references for evaluation, Table \ref{disfluent ref scores} shows the performance of fluent models as evaluated on the original disfluent references. Disfluent target scores are the same as in Table \ref{lit comparison}, and have been copied for easy comparison.
As we would expect, here there is a greater difference in scores. 
The fluent references have fewer long n-gram matches with disfluencies removed, lowering BLEU.
The fluent model's METEOR scores suffer more than BLEU due to the recall calculation; recall on the disfluent references is lower because the fluent model does not produce many of the disfluencies (indeed filler words are not in the vocabulary when trained with the fluent references).
Recall is reduced by ${\sim}14\%$ with the fluent model, %from 62 with the disfluent, 
reflecting the approximate distribution of disfluencies in the original data.
%----------

The differences in scores with these two metrics do not show the full picture. 
Outputs generated by the fluent model are on average 13\% shorter and contain 1.5 fewer tokens per utterance than the disfluent model, which is significant with average utterance lengths of 10-11 tokens.
When scoring the fluent output against the original disfluent references, the shorter length significantly contributes to the lower scores, with the BLEU brevity penalty calculated as 0.86 as opposed to 0.96-1.0 for all other conditions. 
Removing the length penalty from the BLEU score calculation, single-reference scores are boosted to 19.3 and 19.8 from 16.6 and 17.0 for dev and test, respectively (Table \ref{disfluent ref scores}).
This is a somewhat fairer comparison of the disfluent and fluent models, as we do not want the fluent output to match the disfluent sequence length, and the disfluent models are not penalized due to length.
These BLEU scores are now very similar to those of the disfluent model on the disfluent references, though the outputs are very different (Figure \ref{output diff}).
The changes here and the difference in trends between the two metrics with respect to the two types of references show that evaluating this task cannot be simply accomplished with one existing metric: depending on the combination of metric and references, it's possible to mask the difference between disfluent and fluent systems, unless you have word-level disfluency annotations, which are more difficult to obtain.

%---------------------------------------------
\section{Conclusion}
\label{sec:conclusion}

Machine translation applications for speech can suffer due to conversational speech phenomena, particularly the presence of disfluencies.
% Removing disfluencies improves performance of downstream translation, as it causes data to better match typically clean training text.
Previous work to remove disfluencies in speech translation did so as a separate step between speech recognition and machine translation, which is not possible using end-to-end models. 
Using clean references for disfluent data collected by \citet{salesky2018slt}, we extend their text baseline to speech input and provide first results for direct generation of fluent text from noisy disfluent speech. 

While fluent training data enables research on this task with end-to-end models, it is unlikely to have this resource for every corpus and domain and it is expensive to collect. 
In future work, we hope to reduce the dependence on fluent target data during training through decoder pretraining on external non-conversational corpora or multitask learning. 
Further, standard metrics alone do not tell the full story for this task; additional work on evaluation metrics may better demonstrate the differences between such systems.

%-------------------------------------------
\bibliography{min_bib}

\begin{thebibliography}{27}
\expandafter\ifx\csname natexlab\endcsname\relax\def\natexlab#1{#1}\fi

\bibitem[{Bansal et~al.(2018)Bansal, Kamper, Livescu, Lopez, and
  Goldwater}]{bansal2018low}
Sameer Bansal, Herman Kamper, Karen Livescu, Adam Lopez, and Sharon Goldwater.
  2018.
\newblock Low-resource speech-to-text translation.
\newblock \emph{Proc. of Interspeech}.
\newblock ArXiv:1803.09164.

\bibitem[{Burger et~al.(2002)Burger, MacLaren, and Yu}]{burger2002isl}
Susanne Burger, Victoria MacLaren, and Hua Yu. 2002.
\newblock The isl meeting corpus: The impact of meeting type on speech style.

\bibitem[{Chan et~al.(2016)Chan, Jaitly, Le, and Vinyals}]{chan2016listen}
William Chan, Navdeep Jaitly, Quoc Le, and Oriol Vinyals. 2016.
\newblock Listen, attend and spell: A neural network for large vocabulary
  conversational speech recognition.
\newblock In \emph{Acoustics, Speech and Signal Processing (ICASSP), 2016 IEEE
  International Conference on}, pages 4960--4964. IEEE.

\bibitem[{Cho et~al.(2014)Cho, F{\"u}nfer, St{\"u}ker, and
  Waibel}]{cho2014lrec}
Eunah Cho, Sarah F{\"u}nfer, Sebastian St{\"u}ker, and Alex Waibel. 2014.
\newblock A corpus of spontaneous speech in lectures: The kit lecture corpus
  for spoken language processing and translation.

\bibitem[{Cho et~al.(2013)Cho, Ha, and Waibel}]{cho2013crf}
Eunah Cho, Thanh-Le Ha, and Alex Waibel. 2013.
\newblock Crf-based disfluency detection using semantic features for german to
  english spoken language translation.

\bibitem[{Denkowski and Lavie(2014)}]{meteor2014wmt}
Michael Denkowski and Alon Lavie. 2014.
\newblock Meteor universal: Language specific translation evaluation for any
  target language.

\bibitem[{Gal and Ghahramani(2016)}]{gal2016theoretically}
Yarin Gal and Zoubin Ghahramani. 2016.
\newblock A theoretically grounded application of dropout in recurrent neural
  networks.

\bibitem[{Graff et~al.()Graff, Huang, Cartagena, Walker, and
  Cieri}]{ldcfisherspanish}
David Graff, Shudong Huang, Ingrid Cartagena, Kevin Walker, and Christopher
  Cieri.
\newblock Fisher spanish speech ({LDC2010S01}).
\newblock Https://catalog.ldc.upenn.edu/ ldc2010s01.

\bibitem[{Hochreiter and Schmidhuber(1997)}]{Hochreiter1997}
Sepp Hochreiter and J\"{u}rgen Schmidhuber. 1997.
\newblock Long short-term memory.
\newblock \emph{Neural Comput.}

\bibitem[{Honal and Schultz(2005)}]{honal2005spkdisfluencies}
Matthias Honal and Tanja Schultz. 2005.
\newblock Automatic disfluency removal on recognized spontaneous speech-rapid
  adaptation to speaker-dependent disfluencies.

\bibitem[{Kingma and Ba(2015)}]{kingma2014adam}
Diederik~P Kingma and Jimmy Ba. 2015.
\newblock Adam: A method for stochastic optimization.
\newblock \emph{Proc. of ICLR}.
\newblock ArXiv:1412.6980.

\bibitem[{Koehn et~al.(2007)Koehn, Hoang, Birch, Callison-Burch, Federico,
  Bertoldi, Cowan, Shen, Moran, Zens et~al.}]{koehn2007moses}
Philipp Koehn, Hieu Hoang, Alexandra Birch, Chris Callison-Burch, Marcello
  Federico, Nicola Bertoldi, Brooke Cowan, Wade Shen, Christine Moran, Richard
  Zens, et~al. 2007.
\newblock Moses: Open source toolkit for statistical machine translation.

\bibitem[{Kumar et~al.(2014)Kumar, Post, Povey, and Khudanpur}]{kumar2014some}
Gaurav Kumar, Matt Post, Daniel Povey, and Sanjeev Khudanpur. 2014.
\newblock Some insights from translating conversational telephone speech.

\bibitem[{Lardilleux and Lepage(2017)}]{charcut2017}
Adrien Lardilleux and Yves Lepage. 2017.
\newblock Charcut: Human-targeted character-based mt evaluation with loose
  differences.
\newblock \emph{Proc. of IWSLT}.

\bibitem[{Luong et~al.(2015)Luong, Pham, and Manning}]{Luong2015b}
Minh-Thang Luong, Hieu Pham, and Christopher~D. Manning. 2015.
\newblock Effective approaches to attention-based neural machine translation.

\bibitem[{Neubig et~al.(2018)Neubig, Sperber, Wang, Felix, Matthews,
  Padmanabhan, Qi, Sachan, Arthur, Godard et~al.}]{neubig2018xnmt}
Graham Neubig, Matthias Sperber, Xinyi Wang, Matthieu Felix, Austin Matthews,
  Sarguna Padmanabhan, Ye~Qi, Devendra~Singh Sachan, Philip Arthur, Pierre
  Godard, et~al. 2018.
\newblock Xnmt: The extensible neural machine translation toolkit.
\newblock \emph{arXiv:1803.00188}.

\bibitem[{Nguyen and Chiang(2018)}]{nguyen2017improving}
Toan~Q Nguyen and David Chiang. 2018.
\newblock Improving lexical choice in neural machine translation.
\newblock \emph{Proc. of NAACL HLT}.
\newblock ArXiv:1710.01329.

\bibitem[{Papineni et~al.(2002)Papineni, Roukos, Ward, and
  Zhu}]{papineni2002bleu}
Kishore Papineni, Salim Roukos, Todd Ward, and Wei-Jing Zhu. 2002.
\newblock Bleu: a method for automatic evaluation of machine translation.

\bibitem[{Post et~al.(2013)Post, Kumar, Lopez, Karakos, Callison-Burch, and
  Khudanpur}]{post2013improved}
Matt Post, Gaurav Kumar, Adam Lopez, Damianos Karakos, Chris Callison-Burch,
  and Sanjeev Khudanpur. 2013.
\newblock Improved speech-to-text translation with the fisher and callhome
  spanish--english speech translation corpus.

\bibitem[{Povey et~al.(2011)Povey, Ghoshal, Boulianne, Burget, Glembek, Goel,
  Hannemann, Motlicek, Qian, Schwarz et~al.}]{povey2011kaldi}
Daniel Povey, Arnab Ghoshal, Gilles Boulianne, Lukas Burget, Ondrej Glembek,
  Nagendra Goel, Mirko Hannemann, Petr Motlicek, Yanmin Qian, Petr Schwarz,
  et~al. 2011.
\newblock The kaldi speech recognition toolkit.

\bibitem[{Salesky et~al.(2018)Salesky, Burger, Niehues, and
  Waibel}]{salesky2018slt}
Elizabeth Salesky, Susanne Burger, Jan Niehues, and Alex Waibel. 2018.
\newblock Towards fluent translations from disfluent speech.
\newblock \emph{Proc. of SLT}.

\bibitem[{Sperber et~al.(2018)Sperber, Niehues, Neubig, St{\"u}ker, and
  Waibel}]{sperber2018self}
Matthias Sperber, Jan Niehues, Graham Neubig, Sebastian St{\"u}ker, and Alex
  Waibel. 2018.
\newblock Self-attentional acoustic models.
\newblock \emph{Proc. of EMNLP}.
\newblock ArXiv:1803.09519.

\bibitem[{Szegedy et~al.(2016)Szegedy, Vanhoucke, Ioffe, Shlens, and
  Wojna}]{szegedy2016rethinking}
Christian Szegedy, Vincent Vanhoucke, Sergey Ioffe, Jon Shlens, and Zbigniew
  Wojna. 2016.
\newblock Rethinking the inception architecture for computer vision.

\bibitem[{Wang et~al.(2010)Wang, Tur, Zheng, and Ayan}]{wang2010disfluency}
Wen Wang, Gokhan Tur, Jing Zheng, and Necip~Fazil Ayan. 2010.
\newblock Automatic disfluency removal for improving spoken language
  translation.

\bibitem[{Weiss et~al.(2017)Weiss, Chorowski, Jaitly, Wu, and
  Chen}]{weiss2017sequence}
Ron~J Weiss, Jan Chorowski, Navdeep Jaitly, Yonghui Wu, and Zhifeng Chen. 2017.
\newblock Sequence-to-sequence models can directly transcribe foreign speech.
\newblock \emph{arXiv:1703.08581}.

\bibitem[{Zayats et~al.(2016)Zayats, Ostendorf, and
  Hajishirzi}]{zayats2016disfluency}
Vicky Zayats, Mari Ostendorf, and Hannaneh Hajishirzi. 2016.
\newblock Disfluency detection using a bidirectional lstm.
\newblock \emph{Proc. of Interspeech}.
\newblock ArXiv:1604.03209.

\bibitem[{Zhang et~al.(2017)Zhang, Chan, and Jaitly}]{zhang2017very}
Yu~Zhang, William Chan, and Navdeep Jaitly. 2017.
\newblock Very deep convolutional networks for end-to-end speech recognition.

\end{thebibliography}
\bibliographystyle{acl_natbib.bst}

\vspace{1em}
\appendix

\section{Appendix. LSTM/NiN Encoder and Training Procedure Details}
\label{sec:appendix}

\subsection{Encoder Downsampling Procedure}
\citet{weiss2017sequence} and \citet{bansal2018low} use two strided convolutional layers atop three bidirectional long short-term memory (LSTM) \cite{Hochreiter1997} layers to downsample input sequences in \textbf{time} by a total factor of 4.
\citet{weiss2017sequence} additionally downsample \textbf{feature} dimensionality by a factor of 3 using a ConvLSTM layer between their convolutional and LSTM layers. 
This is in contrast to the pyramidal encoder \citep{chan2016listen} from sequence-to-sequence speech recognition, where pairs of consecutive layer outputs are concatenated before being fed to the next layer to halve the number of states between layers. 

To downsample in time we instead use the LSTM/NiN model used in \citet{sperber2018self} and \citet{zhang2017very}, which stacks blocks consisting of an LSTM, a network-in-network (NiN) projection, layer batch normalization and then a ReLU non-linearity. %(Fig. 1b). 
NiN denotes a simple linear projection applied at every timestep, performing downsampling by a factor of 2 by concatenating pairs of adjacent projection inputs.
The LSTM/NiN blocks are extended by a final LSTM layer for a total of three BiLSTM layers with the same total downsampling of 4 as \citet{weiss2017sequence} and \citet{bansal2018low}. 
These blocks give us the benefit of added depth with fewer parameters.

\subsection{Training Procedure}
We follow the training procedure from \citet{sperber2018self}.
The model uses variational recurrent dropout with probability 0.2 and target character dropout with probability 0.1 \cite{gal2016theoretically}.
We apply label smoothing \cite{szegedy2016rethinking} and fix the target embedding norm to 1 \cite{nguyen2017improving}.
For inference, we use a beam size of 15 and length normalization with exponent 1.5.
We set the batch size dynamically depending on the input sequence length such that the average batch size was 36.
We use Adam \cite{kingma2014adam} with initial learning rate of 0.0003, and decay by 0.5 when validation BLEU did not improve first over 10 epochs and after 5 epochs after the first decay.
We do not use L2 weight decay or Gaussian noise, and use a single model replica.
All models use the same preprocessing as previous work on this dataset: lowercasing and removing punctuation aside from apostrophes. 

\end{document}